\documentclass{article}
\usepackage{graphicx} 

\usepackage{booktabs}  
\usepackage{siunitx}   
\title{Fuzzy Hierarchical Complex}
\author{Alexis Kafantaris}

\usepackage[scaled]{helvet}

\usepackage{subcaption}  
\usepackage{tikz}        

\usepackage[utf8]{inputenc}
\usepackage{setspace}
\usepackage{titlesec}
\usepackage{graphicx} 
\usepackage{cite}
\usepackage{amsmath}

\titlespacing{\section}{0pt}{*0}{12pt} 

\usepackage{amsmath, amssymb, amsfonts}
\usepackage{graphicx}
\usepackage{hyperref}
\usepackage{authblk}
\usepackage{geometry}

\title{Fuzzy Hierarchical Multiplex}
\author{Alexis Kafantaris}
\affil{Athens University of Economics and Business}

\date{}

\begin{document}

\maketitle

\begin{abstract}
In this paper a novel and unexplored way to represent systems is analyzed. Extrapolating on the idea of the fuzzy cognitive maps (FCM) and their ability to model causal relationships there is a gap; more precisely, there is no understanding of logical implications \cite{kosko1986fuzzy}. Hence, to achieve that, an alternative perspective is required. Abstracted causality has different mappings for different matters; but, a flat structure that represents a system holistically does not; so, it is suggested that such systems model the state of logical implications of a system. The idea of a fuzzy hierarchical multiplex is exactly that, and also provides a meaningful framework for fuzzy optimization of services.
\end{abstract}

\section{Introduction}
This paper analyzes a fuzzy multiplex from a logical perspective in a way that has not been formalized so far. A fuzzy multiplex is a nested structure with inner nodes representing sub-system level agent traits and with outer nodes representing system agents; all while the ensemble is the system under consideration. Moreover, a mathematical framework is necessary to describe that structure which is formulated and then utilized. The system is firstly initialized using fuzzy set theory\cite{zadeh1965fuzzy}, inspired by Fuzzy Cognitive Maps\cite{kosko1986fuzzy}. Then a criterion that describes the structure is devised to implement a multiplex instead of a map\cite{fernandez2021fuzzy}\cite{karami2020fuzzy}, and lastly system optimization is achieved. Furthermore, the theoretical context behind the multiplex is expounded in an attempt to establish a formal way of handling implications within a closed system using human intelligence. The paper is organized in sections following the reasoning process behind this unique idea.

\section{Conceptual Background}
The idea of hierarchy is to determine which data are relevant for the system and to what degree\cite{zadeh1965fuzzy}; to do so, one needs to define fuzzy relationships and use them for the inputs. These relationships describe the optimization equations of a system, or lets say the causality in general\cite{kosko1986fuzzy}. To address the previous statement, such a system would need an organized structure and might seem like just any other FCM \cite{kosko1986fuzzy}\cite{karami2020fuzzy}. Despite that, it works completely differently and produces results in a way that make sense according to the model's hierarchy of concepts.

The Fuzzy Hierarchical Multiplex (FHM) is a deductive net constructed to measures the sub-system' contribution as well as the systems' outputs. And that is modeled in a way that the system as well as its subsystems can be optimized. A directive for service optimization is given, and the program stems from it\cite{Bai2021}. Something more universal and thorough\cite{Bai2021}\cite{karami2020fuzzy} that can substitute an FCM and even supersede it. 

With advantages and disadvantages compared to the original FCM idea this construct is strictly supervised with respect to each data point; each data points has been used as input for many potential multiplex solutions, so that only an outlier data point perplexes the system. For the multiplex system new equations are derived to determine the constituent update of activation of each node; these equations mimic a simple FCM update but are slightly more thorough.

$$\text{ODE}
$$
$\text{Do: } $
$$\nabla (g(\sigma(yW + x)) =$$

$\text{Do: }$
$$\nabla (uV+z) = (u+z)(u+1) + 2z + 1$$

$$\frac{\partial (yW + x)}{\partial u} + \frac{\partial (yW + x)}{\partial v} + \frac{\partial (yW + x)}{\partial z}$$

$\text{Let: }$ 
$$ g(\sigma(yW+x)) = g(yW + x) = f(uV+z) = f(u,V,z)
$$

$$= \frac{\partial f(uV+z)}{\partial u} + \frac{\partial f(uV+z)}{\partial v} + 
\frac{\partial f(uV+z)}{\partial z}
$$

$$= (u+z) + (u+z) + uv +2z + 1 = 1 $$
$$ u v + u v + 2 z + L = L $$
$$ = (v+1)(u+1) + 2z - L \quad (2)$$
$$ Where \ L \ the \ data \ Targets.$$\newline

To determine the implication on the multiplex one has to fit the input data to targets, and one way to achieve that is by minimizing the L2 norm. Moreover, it seemed that by maximizing the fit using the aforementioned ODE captured the interconnections of assigned data to given metrics. There was neither a condition nor any fancy approach, yet, forcing the multiplex to minimize L2 norm to data capture the hierarchy of the system metrics.

In other words, one imitated the input data and solved for L2 fit using the gradient descent. This way, the system was assumed to work optimally according to its initial state. Furthermore to simplify the process, as the data had to fit a multiplex correctly with variable target metrics, an ordinary differential equation was devised based the hierarchy of the multiplex; this ODE substituted the sigmoid of x,y,W for z,u,V and provided always a solution.

The ODE that was described above is designed to fit the data on any basis to the multiplex structure; this way a tensor is formed that always points to nodes and input relations of the multiplex. The tensor can then be stated to capture the implication logic of the system as well as hierarchy and will be examined more thoroughly upon interpretation section.

\section{Mathematical Structure of the Multiplex}
Few words about the model; the FHM model is a bidirectional soft targeting hierarchical fuzzy multiplex. The optimal state of the model is achieved when it learns the data for some quantifiable objectives. In simple words, the data is not just memorized but resolved according to metric criteria. To achieve that the outer layer is in place constraining the inner layer activations to optimal patterns. In turn that allows for the context of the hierarchical structure can be captured by the FHM model; here is the simplest form of its mathematical formulation. 

\[
\begin{aligned}
z_i^{(t+1)} &= f\big(z_i^{(t)}, Y^{(t)}\big) \\
Y^{(t+1)} &= w\big(Y^{(t)}, \{z_i^{(t+1)}\}\big)
\end{aligned}
\]

- \( f \): Inner multiplex update function 

- \( w \): Outer multiplex objective function

- \( z_i \): conceptual or subsystem semi-ordered inner nodes

- \( Y \): integrative or cognitive outer nodes or system problems

Additionally, the multiplex optimizes objectives simultaneously in the inner layer and at the outer, based on these criteria. During the optimization process the inner layer structures act independently and change dynamically with the outer nodes, yet no other fuzzy or logical constrain is imposed; the evolution of each inner sub-system is evaluated on a local level and is measured based on information metrics. The programmed function is matching metric quantities of data fitness according to the multiplex. That resolves a great deal of issues, both theoretical and practical. For example there is no overfit of data metric, while there is plenty of meaning in fine tuning the structure to produce higher throughput for a working facility\cite{Bai2021}\cite{KAHRAMAN2006390} Finally, the model creates multiple instances of the targeted data or soft solution. An inter layer for the dimension is implied and can be also implemented.

\begin{figure}[h!]
    \centering
    \includegraphics[width=0.7\textwidth]{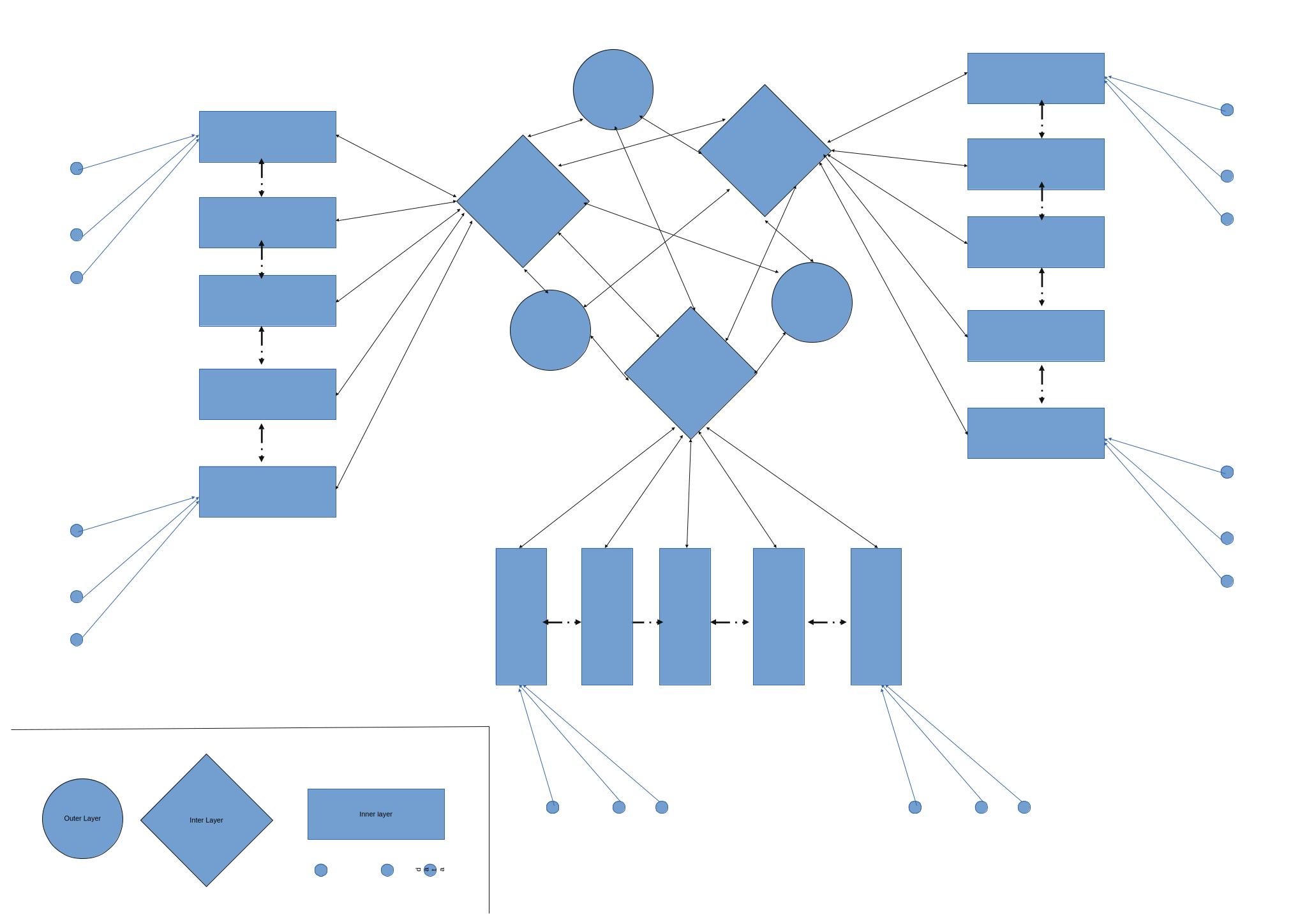}  
    
    \caption{Prototype Multiplex with inter layer}
    \label{fig:Fuzzy Hierarchical multiplex Sketch}
\end{figure}\pagebreak

\section{Interpretation}

Nevertheless, extended from the fuzzy cognitive maps to the fuzzy hierarchical multiplex a rough sketch is displayed above. Moreover, the FHM is a weighed structure that describes a system, its sub-systems, and their interconnection for a given state;  overfiting the data won t cut it, while the metrics on their own are not a part of any system. For the multiplex any system is adapting the data to map metrics within fuzzy intervals. Let it be visualized in the 3D space as a raw data structure. Starting from the outer layer, the outer layer is a essentially a graph with nodes that contain nested FCMs; it is modeled by systemic global objectives, like information signal and comprised from a nested inner layer of FCMs according to the problem specifications. Its physical interpretation is that of a system that contains sub-systems on a node level; note that it can also factor inter layer interactions and optimize them through multiplexing.

\begin{figure}[h!]
    \centering
    \includegraphics[width=0.7\textwidth]{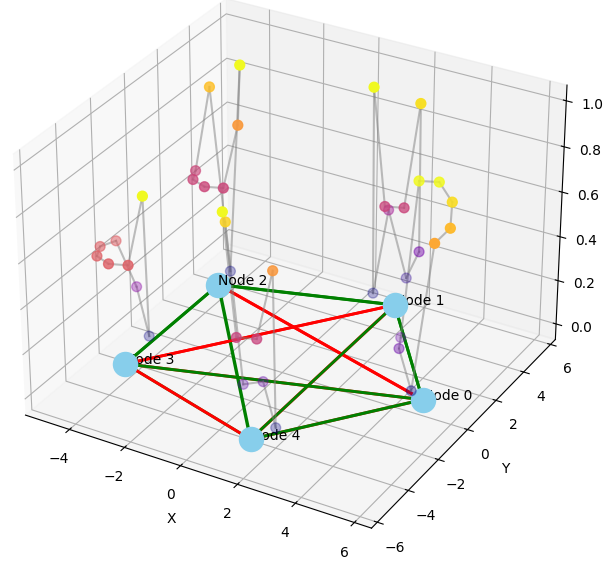}  
    
    \caption{Prototype Hierarchical Multiplex}
    \label{fig:Fuzzy Hierarchical multiplex}
\end{figure}\newpage

That achieves interpret ability of system behavior attributes when the system is examined according to data metrics, e.g. a bottleneck in throughput hinders the overall performance of the entire system. For this reason the metrics transcend the operations and an effort is made to align them with the data in a holistic manner; provided a specific activation for given metric targets instead of is totaly explanatory of implication.\newline

\begin{figure}[h!]
    \centering
    \begin{tikzpicture}[baseline=(current bounding box.center)]
        \node (img1) at (0,0) {
            \includegraphics[width=0.35\textwidth]{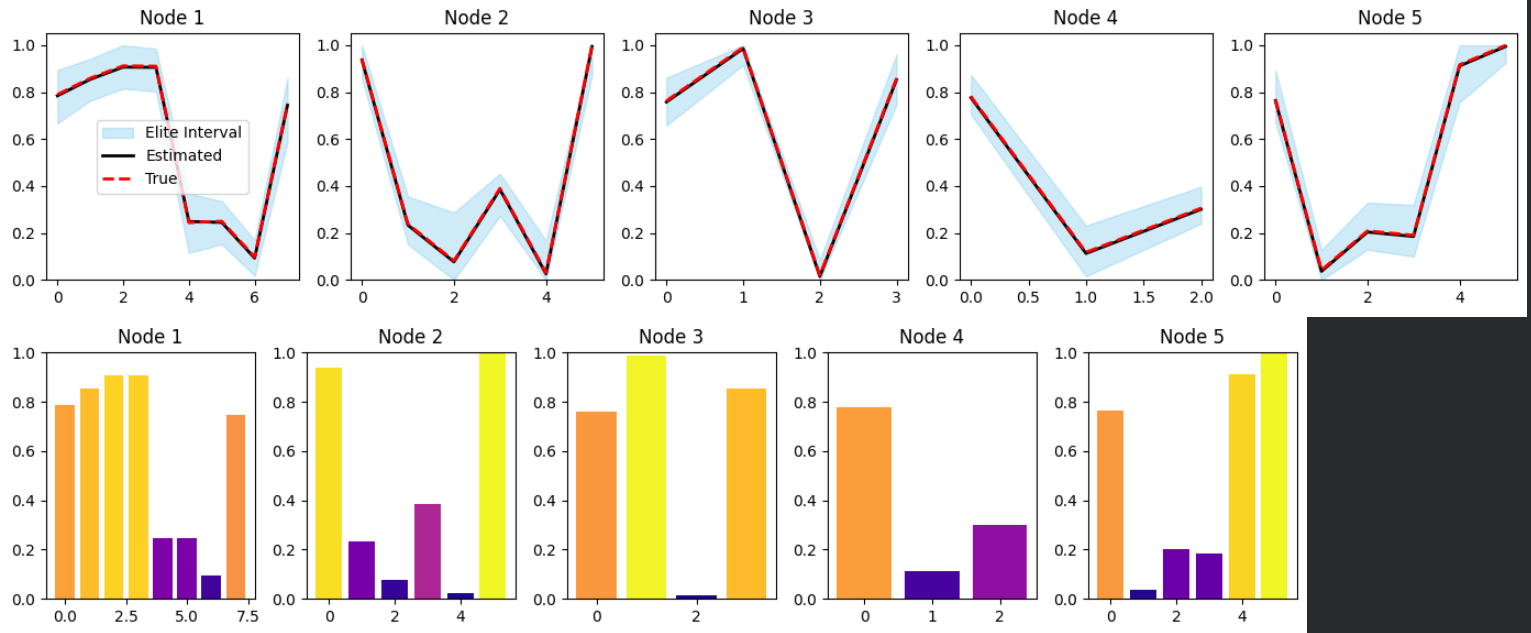}
        };
        \node (img2) at (6,0) {   
            \includegraphics[width=0.35\textwidth]{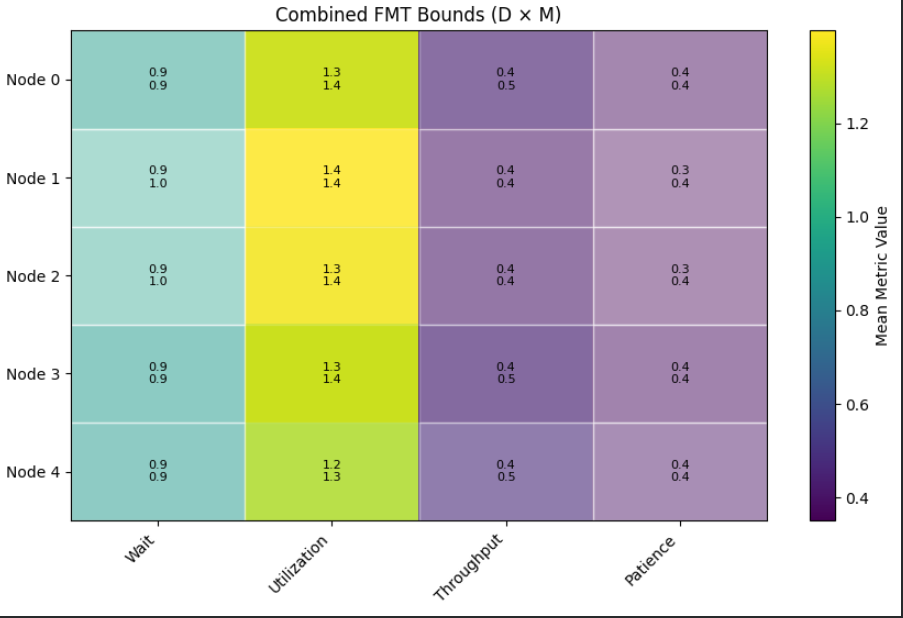}
        };
        \draw[->, thick, >=stealth] (img1.east) -- (img2.west);
    \end{tikzpicture}

    \caption{Data Multiplex (left) transforming into Metric Multiplex (right).}
    \label{fig:arrow_images}
\end{figure}
The hierarchy lies in the activations as from given inputs the current system state is represented as displayed above. On the left side of figure 3 the inner activations are learned and on the right side the given metrics they yield are calculated for each node.  In this example, an inter layer has been assumed and interconnections are also affecting the outcomes. A generic example of the\cite{chen2020hims} in a well established framework is illustrated. Here the same parameters were factored, albeit the approach was completely different.

Moreover, it can be attested that the system is deterministic and produces results that match one to one with the inputs. The inputs are solved either heuristically or using exact methods for a given problem, problems such as \cite{chen2020hims}\cite{karami2020fuzzy}, and provide also deterministic output. That means, calculated metrics from the data are one to one also. If the input data is fuzzy, the output metrics and relationships will also have a degree of uncertainty\cite{kosko1986fuzzy}\cite{zadeh1965fuzzy}. Lastly the mappings of the metrics which are in this case utility, patience, wait, and throughput are concomitantly solved heuristically or exactly and in parallel to provide an overall solution as well as an optimized multiplex.

Taking a step backward determining the mappings as illustrated comes from a multiplex and even if one were to use ones solution target, would result in an Interval valued Fuzzy Set (IVFS) for the metrics due to the nature of the systems that calculate these relationships. The system does try to map the solutions exactly however there are potential mappings that may vary upon occasion. Nonetheless, the exact way in which the system is created has been formalized and established; and so to the hierarchical evaluation of nodes will be shown. Meanwhile, predicting the per node metrics is not a new concept and neither correlating the metrics with data is anything special. What gives meaningful correlation is the ability of the system to align its predictions with the data on a multiplex level, and hence, the hierarchical structure; the transcendence of data into a meaningful hierarchical structure is what matters after all.

\begin{table}[h!]
\centering
\caption{Evaluation Metrics and Node Contributions}
\label{tab:gen9_metrics}
\sisetup{
  round-mode=places,
  round-precision=2,
  table-format=1.2
}
\begin{tabular}{c S S S S S}
\toprule
Node & {Wait} & {Throughput} & {Utilization} & {Patience} & {Contribution} \\
\midrule
0 & 0.69 & 0.73 & 0.69 & 0.70 & 7.29 \\
1 & 0.70 & 0.73 & 0.70 & 0.72 & 7.19 \\
2 & 0.72 & 0.68 & 0.69 & 0.71 & 5.83 \\
3 & 0.72 & 0.69 & 0.71 & 0.67 & 6.11 \\
4 & 0.70 & 0.71 & 0.71 & 0.71 & 5.91 \\
\bottomrule
\end{tabular}
\end{table}
Notice how the nodes can be sorted by contribution in table 1 and that the complete dynamics of the problem have been resolved in the apparatus of figure 2,3. A note here, each node represents a different sub-system of the total.

\section{Practical applications}
This FHM produces an IV fuzzy set that is meant to model behavioral logic as a collection of inner patterns that adhere optimally to the overall system. It is noteworthy, that the practical basis was derived from service optimization\cite{Bai2021}; a service design optimizer perceives a service as an abstract network of information with nodes described as processes. These processes in turn are modulated and then correlated in order to address implications the best way possible by using a criterion. In other words, the main objective is to transcend the data into information and the into knowledge of the system. And that case is evident in many of the examples; in an attempt to model services as procedures\cite{chen2020hims}\cite{Bowie2009} to, a generalization for the FHM was derived. 

Or in other words, to accomplish the aforementioned objective, one had to create a framework in which a multiplex can be meaningfully used as an optimizer. Furthermore different synthetic data are evaluate, such as data corresponding to problems like multi-objective multi criteria\cite{chen2020hims} or grid optimization\cite{karami2020fuzzy}, using the FHM.  It is attested that produces good outputs, ultimately aligned with the data and not so much the dynamics. In comparison with the FCM that is, the FHM always excels at minimizing the errors and aligning with the data which is to say, it derives the implications based on patterms; it uses metric-data alignment to determine overall system patterns, a mind blowing idea that is. 

\section{Conclusion}
To conclude, the fuzzy hierarchical multiplex is a new type of model that comes with a purpose; the multiplex is meant to solve multi-purpose and multi-objective problems and other kind of problems which are not described or referenced here. It addresses and resolves some of the service design problems to, which are of interest\cite{Bai2021} and may have further applications in various fields of optimization to say the least\cite{kosko1986fuzzy}\cite{drakopoulos2020tensorfcm}. It is asserted that there are more things that it can do, such as predict-generate relevant.

Nevertheless, a fuzzy hierarchical multiplex is developed and presented because it is a new idea. This multiplex was designed to model service processes in the context of fuzzy optimization; however, that development was generalized into a new structure that displays some very interesting properties especially when optimized with neural nets... The most interesting one is its ability to adapt to current metrics in a way that a global as well as a local set of objectives is achieved optimally and holistically. That is to say it models the system in a new and interesting way. 

Lastly, it may resemble an FCM variant but it is assured that it works in a completely different manner. It is a more direct approach and one that provides best alignment with the objectives at a systems' level. It models patterns within the system and its subsystems giving a thorough instigation on the specifics throughout the different levels of hierarchy optimization. Hence, it is suggested as a fuzzy hierarchical multiplex or FHM, and it is very promising.

\pagebreak

\section*{Acknowledgments}
It is acknowledged that this paper is part of a PhD dissertation, fuzzy optimization of information transmission in service design process currently done in (Athens University of Economics and Business) AUEB. It is also written in correspondence with Dr. Dimitris Kardaras which was the supervisor of the specific subject in AUEB and also interested in service optimization.

\end{document}